# A Survey on Deep Industrial Transfer Learning in Fault Prognostics

Benjamin Maschler[1]

**ABSTRACT**

Due to its probabilistic nature, fault prognostics is a prime example of a use case for deep learning utilizing big data. However, the low availability of such data sets combined with the high effort of fitting, parameterizing and evaluating complex learning algorithms to the heterogenous and dynamic settings typical for industrial applications oftentimes prevents the practical application of this approach. Automatic adaptation to new or dynamically changing fault prognostics scenarios can be achieved using transfer learning or continual learning methods. In this paper, a first survey of such approaches is carried out, aiming at establishing best practices for future research in this field. It is shown that the field is lacking common benchmarks to robustly compare results and facilitate scientific progress. Therefore, the data sets utilized in these publications are surveyed as well in order to identify suitable candidates for such benchmark scenarios.

## 1. INTRODUCTION

Fault prognostics is the ability to predict the time at which an entity becomes dysfunctional, i.e. faulty [1]. Depending on the entity and its environment, causes for faults can be diverse and complex, rendering fault prognostics highly probabilistic. In recent years, the combination of big data and deep learning methods have demonstrated to have a great potential [1–3]. However, many approaches delivering promising results in research environments fail to achieve wide-spread utilization in industry [4–7].

**One major challenge** towards a wider adoption is the high effort of fitting, parameterizing and evaluating complex learning algorithms to the heterogenous and dynamic settings typical for industrial applications [1, 3, 4]. This is especially severe for fault prognostics, as failures are usually to be avoided in productive systems, making the collection of labeled training data sets even harder than usual. A lack of automatic adaptability and ready-to-use architectures or frameworks diminishes the benefits data-driven fault prognostics could bring and thereby deters potential users [4].

It is therefore necessary to lower the effort of adapting learning algorithms to changing problems – may those changes be caused by internal problem dynamics or by the problem being new and dissimilar from others. **One promising approach** is the transfer of knowledge between different, unidentical instances of a problem, e.g. by deep transfer or deep continual learning [4]. However, this being a recent research trend, not much comparison or benchmarking of methods nor results has been published and no best practice analysis been performed.

Although there are first surveys on transfer learning in technical applications in general [8], there are none on fault prognostics in particular, yet. Therefore, the **objective of this article** is

- to provide a brief introduction in industrial transfer learning methods as well as fault prognostics basics in order to facilitate mutual understanding between experts of the respective fields,

[1] University of Stuttgart, Department of Computer Science, Electrical Engineering and Information Technology, Pfaffenwaldring 47, 70569 Stuttgart, Germany, +49 711 685 67295, benjamin.maschler@ias.uni-stuttgart.de, ORCID: 0000-0001-6539-3173

- to provide a comprehensive overview of published research activities in the field of deep industrial transfer learning for fault prognostics and to analyze it with regards to best practices and lessons learned in order to consolidate scientific progress and, thereby, offer guidance for future research projects,
- to provide a comprehensive overview of open-access fault prognostics data sets suitable for deep industrial transfer learning research and to analyze it in order to lower the threshold for new research in this field and allow for a benchmarking of results.

This **article is organized** as follows: Chapter 2 introduces the concepts and methods of deep industrial transfer learning as well as the basic principles of fault prognostics. Chapter 3 then briefly describes the research methodology. Chapter 4 is divided into two parts: The first part presents the surveyed publications while the second part presents the corresponding data sets. Chapter 5 retains this structure, discussing first the publications and then the data sets. Chapter 6 concludes this article and points out new research directions.

## 2. RELATED WORK

In this chapter, first, the concepts and methods of deep industrial transfer learning are introduced. Then, an overview of the principles of fault prognostics is presented. Both serve to set the terminology for the remainder of this article and facilitate understanding between experts of the respective fields.

### 2.1 DEEP INDUSTRIAL TRANSFER LEARNING

A general approach to overcome the described challenges is the transfer of knowledge across multiple (sub-) problems. This makes it possible to create a (more) complete model of the problem to be solved across different scenarios and to adapt it dynamically again and again without the need to completely retrain the algorithm representing said model every time.

In machine learning research, a distinction is made between two families of solutions: **transfer learning**, which aims solely at better solving a new target problem [9], and **continual learning**, which aims at solving a new target problem while maintaining the ability to solve previously encountered source problems [10]. In practice, however, this theoretical division often turns out to be unsuitable, since both the generalization capabilities of continual learning and the specialization capabilities of transfer learning might be helpful in extracting generalities from various, known (sub-)problems and then adapting them to the problem at hand [8]. This is to be represented by the term **industrial transfer learning** [4, 8, 11].

Different methods of machine learning can be utilized in the context of (industrial) transfer resp. continual learning, e.g. artificial neural networks, support vector machines or Bayesian networks [9, 12]. When only deep learning methods are used, this is referred to as **deep industrial transfer learning**.

If the transfer of knowledge from known problems can influence the learning of solutions to new problems, then such an influence does not necessarily have to be positive. A harmful knowledge transfer is therefore called **negative transfer** [9, 12]. It is defined with regards to the difference in performance of a given learning algorithm with and without using data of the source problem, the so-called transfer performance. If the performance is better without using data of the source problem, negative transfer is present.

In practice, there are three main approach categories used in deep industrial transfer learning. The following sub-chapters will introduce them briefly.

#### 2.1.1 Feature representation transfer

**Feature representation transfer** includes approaches which map the samples from the source and target problems into a common feature space to improve training of the target problem [8, 9, 12]. A central concept of feature representation transfer is **domain adaptation** [9, 12]. A distinction is made between unsupervised domain adaptation, i.e., requiring no target labels at all, and semi-supervised domain adaptation, i.e., requiring only a few target labels. Domain adaptation is usually based on minimizing the distance between the feature distributions of the different (sub)problems. Common metrics for this distance are the maximum mean discrepancy (MMD) or the Kullback-Leibler (KL) divergence.

#### 2.1.2 Parameter transfer

**Parameter transfer** includes approaches, which pass parameters or initializations from the learning algorithm trained on the source problem to the target problem learning algorithm to improve its initialization before actual training on the target problem begins [8, 9, 12]. Two forms of parameter transfer can be distinguished in deep industrial transfer learning:

Partial parameter transfer includes approaches that pass only the parameters of the feature extractor from the learning algorithm trained on the source problem to the target problem learning algorithm [13]. The feature extractor is then not subject to the training on the target problem but remains static throughout that phase. Such use of a **shared feature extractor** reduces the training effort on the target problem, because only a small part of the entire learning algorithm still needs to be trained [14].

Full parameter transfer includes approaches that pass all parameters and initializations of the learning algorithm trained on the source problem to the target problem learning algorithm [13]. The transferred learning algorithm is then further trained on the target problem. This process, also called **finetuning**, reduces the training effort on the target problem, because the learning algorithm is already pre-trained [14].

*2.1.3 Regularization*

**Regularization**-based continual learning includes approaches which extend the loss function of an algorithm to penalize the changing of parameters that were important for solving previously learned tasks [8, 10]. It is therefore related to finetuning, because it involves passing all parameters and initializations of the learning algorithm trained on the source problem to the target problem learning algorithm. However, most of the prominent examples of regularization methods are only suitable for classification problems [15].

*2.2 FAULT PROGNOSTICS*

A fault is understood to be the arrival at a state of dysfunctionality. In relation to industrial components, this can be accompanied by the failure of other components or higher-level systems. Due to the high costs associated with unplanned failures, fault prognostics is an important field of research. Its subject is the prediction of the timing of a failure - usually without further consideration of its cause. Its goal is, among other things, to enable proactive maintenance, to increase operational safety and to reduce fault costs [2, 16–18].

Usually, three different approaches to fault prognostics are distinguished:

**Model-based approaches**, further subdivided into physics-based and expert-based approaches [17], describe deterioration processes using mathematical models and starting from the causes of faults and the factors influencing them. Such approaches are very efficient and accurate but require a complete understanding of the system. Adaptation to new or changed scenarios usually has to be done manually and is therefore only worthwhile for static and expensive scenarios [2, 16]. **Data-based approaches**, further subdivided into numerical or statistical and machine learning-based approaches [17], describe deterioration processes on the basis of historical data, which can be obtained either from the real plants, from test setups or simulations. Such approaches are usually inexpensive to develop, easy to adapt, and require little or no understanding of the system. On the other hand, they are very demanding in terms of variety, quantity and quality of data, and sometimes require a lot of training [2, 16]. **Hybrid approaches** combine model- and data-based methods with the goal to increase the quality of data-based approaches without generating the high effort of exclusively model-based approaches [2, 16].

There is a large canon of research on data-based fault prognostics using deep learning methods. They can be categorized into two different classes based on their objective:

The prediction of the **Remaining Useful Lifetime** (RUL) as a continuous percentage of the total lifetime is the usual approach for fault prognostics [1, 3, 19, 20]. It is a regression problem. CNN, RNN, Autoencoder and Deep Belief Networks are mainly used [1–3]. Occasionally, time series prediction using LSTM is also used for RUL prediction [21]. Due to the fact that many characteristics due not significantly change during the first part of the total lifetime [22], sometimes a piecewise linear RUL (p-RUL) is used instead of the fully linear RUL.

An alternative to RUL prediction is **State of Health** (SoH) estimation in the form of a multi-class classification. While the term SoH originally originated in the field of battery research, where it was also a continuous percentage $SoH_\%$ related to the ratio of different actual performance parameters to their respective nominal values [19, 20, 23], the term is now also used for a discrete classification of the remaining lifetime $SoH_{class}$ - both in the battery context [24, 25] and beyond [26–29]. $SoH_{class}$ estimation is primarily used when RUL prediction is not possible, for example, for methodological reasons or due to insufficient training data. In some cases, the more accurate RUL prediction is deliberately omitted because the $SoH_{class}$ estimation is less complex and sufficient for the application at hand. Occasionally, the $SoH_{class}$ is also used as a preliminary stage of a subsequent RUL prediction [30, 31].

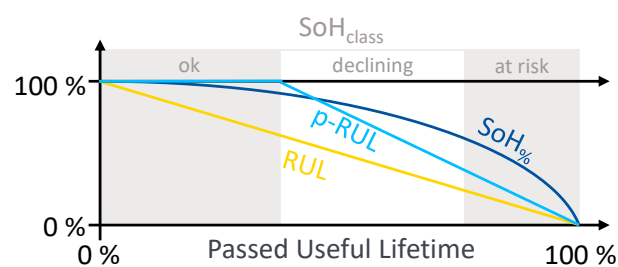

FIGURE 1.   **RUL, p-RUL, SoH% and SoHclass as a function of useful lifetime already passed**

Figure 1 shows an example of RUL, p-RUL, $SoH_{\%}$ and $SoH_{class}$ as a function of useful lifetime already passed. It can be seen that $SoH_{\%}$ is advantageous over linear RUL in the context of strongly non-linear behavior of e.g. battery capacity [19, 20] - however, in other areas with a more linear behavior or a focus on the remaining lifetime as opposed to the remaining functionality, it does not provide any advantage. Subsequently, SoH will therefore be used in the sense of $SoH_{class}$ throughout this article.

## 3. RESEARCH METHODOLOGY

In order to present a comprehensive overview of the current state of research in the field of deep industrial transfer learning for fault prognostics, in this study, a two-stepped systematic literature review is conducted.

In the **first step**, publications matching a combination of search terms are selected on Google Scholar. The search terms are listed in Table 1. One term of the category "term 1" and one term of the category "term 2" was selected for each search query.

In the **second step**, a manual selection process further filtered those publications. Only full-text, English language, original research publications that utilized deep learning methods and some kind of transfer or continual learning on fault prognostics use cases and were published until the end of 2021 were to be included in the detailed analysis for this study.

**TABLE 1.**    Search terms

| *Term 1* | *Term 2* ||
|---|---|---|
| Transfer Learning | Fault Prognostics | Fault Prognosis |
| Continual Learning | Remaining Useful Lifetime | State of Health |

## 4. RESULTS

In this chapter, the results of the systematic literature review are presented. First, the original research publications, the methods and scenarios utilized therein are described. Then, open-access fault prognostics data sets suitable for deep industrial transfer learning are introduced.

### *4.1 APPROACHES*

The publications matching the criteria described in section 3 are listed in TABLE 2. All of them are studies involving deep-learning-based transfer or continual learning on fault prognostics use cases. In the following, they are analyzed grouped by their transfer approach category.

[32] uses **domain-adaptation** for **semi-supervised** RUL prediction. The described prototype combines long short-term memory (LSTM) as feature extractor, an unspecified neural network as discriminator and fully connected neural networks (FCNN) as regressor. An evaluation on the NASA Turbofan data set demonstrates the positive transfer between source and target problems. The approach without transfer functionality was trained on the source problem only, because no labels should be necessary for training the target problem. Comparisons with other domain adaptation algorithms from [33] also revealed better performance of the presented algorithm. The prototype described in [34] uses a combination of convolutional neural networks (CNN) as feature extractor, FCNN as discriminator and FCNN as regressor. An evaluation on the NASA Milling data set demonstrates the positive transfer between source and target problem, especially when only a small number of target samples is used. However, because only a single transfer scenario is considered, the validity of the study is limited.

[35] uses **domain-adaptation** for **unsupervised** RUL prediction. The described prototype combines stacked denoising autoencoders as domain adaptor and FCNN as regressor, where only the feature extractor is adapted to the target problem and the regressor remains fixed. Thus, an unsupervised adaptation of the supervised pre-trained algorithm to the target problem is possible. An evaluation on a proprietary, univariate milling data set consisting of vibration data demonstrates the positive transfer between source and target problems. The prototype described in [33] uses a combination of LSTM as feature extractor and FCNN as regressor and discriminator. An evaluation on the NASA Turbofan data set demonstrates the positive transfer between source and target problems, utilizing a separate hyperparameter optimization for each transfer scenario. A comparison with other unsupervised domain adaptation algorithms (including [36]) shows that the presented algorithm achieves higher accuracy and, additionally, a comparison with the supervised approach of [37] is also in its favor. Furthermore, the alternative use of FCNN, CNN or recurrent neural networks (RNN) as feature extractors is investigated, with the best overall

results obtained on an LSTM basis. In contrast, [30] uses a combination of FCNN and bidirectional gated recurrent units (GRU) as regressors. These are preceded by a feature extraction, which generates previously defined features from the time series signals, which are then checked for their domain invariance before further use. An evaluation on FEMTO-ST Bearing data set demonstrates the positive transfer between source and target problems. A comparison with other algorithms (among others [36, 38]) shows that the presented algorithm achieves a higher accuracy and that the special feature extraction heavily contributes to this. [38] describes a combination of CNN and FCNN as feature extractors, multi-kernel MMD as objective function for the adaption process and FCNN as regressor. An extensive evaluation is performed on the FEMTO-SE Bearing data set which, however, is only used in a univariate fashion after a fast Fourier transformation (FFT): At first, only considering the described prototype, the optimal parametrization of the objective function is investigated. It is shown that the MMD should be determined on features extracted as late as possible, i e. on the results of the feature extraction. By means of a comparison with learning algorithms of the same architecture but without transfer functionality and trained on labeled source, target or source and target samples, it is shown that the presented algorithm produces a positive transfer and also generalizes better. A final comparison with other algorithms, which however only sporadically use deep transfer learning, demonstrates that the presented algorithm achieves higher accuracy. In [39], the authors of [38] present two other approaches to unsupervised RUL prediction using domain adaptation - an adversarial approach and a non-adversarial one. Both prototypes described use a combination of CNN as feature extractor and FCNN as regressor. The non-adversarial approach is based on an extended loss function that considers the marginal probability distribution using multi-kernel MMD and the conditional probability distribution based on a fuzzy class division (analogous to SoH classes) in combination with MMD. The adversarial approach uses a modular discriminator for the marginal and conditional probability distributions, although their internal structures are not described in detail. Via a reverse validation based source sample selection, suitable source samples are identified before starting the actual adaptation process. On the XJTU-SY and FEMTO-ST Bearing data sets, a comprehensive evaluation is performed – again, on univariate time series of the FFT'ed raw data: By means of comparisons with learning algorithms of different architectures with and without transfer functionality and different training strategies, it is demonstrated that both algorithms presented produce a positive transfer as well as perform (in many cases significantly) better. On one data set, the non-adversarial approach comes out ahead, on the other the adversarial approach. The prototype described in [40] uses a combination of FCNN as feature extractor, multi-kernel MMD as objective function for the adaptation process and kernel regression as regressor. An evaluation on FEMTO-ST data set demonstrates a positive transfer between source and target problem, although the overall predictive accuracy is rather low. Instead of using raw data, the spectral power density (PSD) is used as an input. A comparison with other algorithms (including [33] and [38]) shows that the presented algorithm achieves higher accuracy. However, it remains unclear on which numbers this comparison is based and the average values given for the presented approach are not taken from the publication itself. [41] describes a combination of CNN as feature extractor, FCNN as discriminator and FCNN as regressor. An evaluation on the FEMTO-ST data set demonstrates the better prediction quality of the algorithm compared to several other algorithms (among others [38]). Even though these other algorithms include one without transfer functionality, a direct evaluation of the transfer performance is not possible due to its different architecture. Instead, an investigation of the effect of different learning rates and kernel sizes is performed. Concludingly, [42] uses a combination of CNN and FCNN as feature extractors and FCNN as regressors. Via extensions of the loss functions for "healthy" and failure-prone samples (similar to SoH classes), a separate domain adaptation is performed for these two categories respectively. An evaluation on the NASA Turbofan and XJTU-SY Bearing data sets demonstrates a positive transfer between source and target problems. The influence of the transfer functionality is investigated for a conventional domain adaptation as well as for the proposed domain adaptation with separate treatment for different SoH classes. The proposed approach achieves the best results, but also requires a longer training time.

[37] uses **finetuning** for supervised RUL prediction. The described prototype combines bidirectional LSTM as feature extractor and FCNN as regressor. An evaluation on the NASA Turbofan data set demonstrates the positive transfer between the source and target problems. It is observed that even in the case of larger differences between source and target problems, e.g., in terms of the number of operating or fault conditions, the transfer was mostly positive. [43] combines pre-trained CNN as feature extractor and own bidirectional LSTM and FCNN as regressor. An evaluation on the XYTU-SY Bearing data set and a Proprietary Gearbox data set, whose univariate time series are, however, both converted to images, demonstrates a positive transfer between the generic ImageNet data set as source problem and the aforementioned target problems. Moreover, the training time with transfer functionality is smaller than without. A comparison with other algorithms shows that the presented algorithm achieves higher accuracy. However, it remains unclear why the feature extractor is not adapted more to the present use case - for example, regarding the same image being processed three times because the pre-trained algorithm used has three input channels. Moreover, [44] compares different approaches of **parameter transfer** for supervised RUL

prediction. Based on an investigation of different deep learning methods without transfer functionality, a prototype combines LSTM and FCNN as regressors is described. An evaluation on a proprietary air compressor data set and the NASA Turbofan data set demonstrates a positive transfer between the source and target problems. Here, different approaches to parameter transfers, e. g. of sub-algorithms that can be finetuned as well as sub-algorithms whose parameters were kept static, were investigated.

[45] uses a **shared feature extractor with finetuning** for supervised RUL prediction. The described prototype combines LSTM as feature extractor and FCNN as regressor, preserving the feature extractor without any changes and adapting only the regressor to the target problem, if necessary. This adaptation depends on the result of the Gray Relational Analysis [46] of handcrafted features of specific univariate time series of the source and target data set. An evaluation on the NASA and CALCE Battery data sets proves the positive transfer between source and target problems especially related to the period just before failure occurrence. Furthermore, the training time with transfer functionality is smaller than without. A comparison with other algorithms shows that the presented algorithm achieves a higher accuracy than other deep-learning-based algorithms but a lower accuracy than other non-deep-learning based algorithms. [47] also uses a shared feature extractor with finetuning, here in the form of a reduction of the MMD between the probability distributions of the source and target problems' extended loss function, for supervised, indirect RUL prediction. Specifically, the wear of the cutting edge of cutting tools is determined directly, which is supposed to allow the indirect inference of the RUL. The described prototype combines a pre-trained CNN as feature extractor and a custom FCNN as regressor. Only the regressor is adapted to the target problem and the feature extractor remains unchanged. An evaluation on an industrial image data set returns a high accuracy value but does not provide any comparative results. Thus, neither an assessment of the relative performance nor of the transfer performance is possible.

[27] uses **regularization-based continual learning** for supervised SoH estimation. The described prototype combines LSTM and FCNN as classifier. An evaluation on the NASA Turbofan data set demonstrates a positive, even multiple transfer between source and target problems. A strong dependence of the transfer performance on the similarity of source and target problems is described. [25] expands on those findings, using a similar architecture on the NASA Battery data set. Again, a positive, multiple transfer between source and target problems can be shown. However, the strong dependence of transfer performance on similarity, direction, sequence, and number of source and target problems, which has not yet been investigated in detail, is described as problematic as it naturally greatly influences the approach's applicability.

**TABLE 2.    Overview of publications utilizing deep industrial transfer learning for fault prognostics**

| Source | Learning Category | Problem Category | Data Type(s) | Data Set(s) | Transfer Category |
|---|---|---|---|---|---|
| Zhang et al. (2018) [37] | Supervised | RUL | Multivar. Time Series | NASA Turbofan | Finetuning |
| Sun et al. (2019) [35] | Unsupervised | RUL | Univar. Time Series | Proprietary Milling | Domain Adaptation |
| Da Costa et al. (2020) [33] | Unsupervised | RUL | Multivar. Time Series | NASA Turbofan | Domain Adaptation |
| Maschler et al. (2020) [27] | Supervised | $SOH_{class}$ | Multivar. Time Series | NASA Turbofan | Regularization |
| Ragab et al. (2020) [32] | Unsupervised | RUL | Multivar. Time Series | NASA Turbofan | Domain Adaptation |
| Russell et al. (2020) [34] | Semi-supervised | RUL | Univar. Time Series | NASA Milling | Domain Adaptation |
| Tan et al. (2020) [45] | Supervised | $SOH_{\%}$ | (Hand-crafted) Features | NASA Battery; CALCE Battery | Shared Feature Extractor plus Finetuning |
| Zhang et al. (2020) [43] | Supervised | RUL | (Self-generated) Images | ImageNet; XJTU-SY Bearing; Proprietary Gearbox | Finetuning |
| Cao et al. (2021) [30] | Unsupervised | RUL | Univar. Time Series | FEMTO-ST Bearing | Domain Adaptation |
| Cheng et al. (2021) [38] | Unsupervised | RUL | Univar. Time Series | FEMTO-ST Bearing (FFT) | Domain Adaptation |
| Cheng et al. (2021) [39] | Unsupervised | RUL | Univar. Time Series | XJTU-SY Bearing (FFT); FEMTO-ST Bearing (FFT) | Domain Adaptation |
| Ding et al. (2021) [40] | Unsupervised | RUL | Univar. Time Series | FEMTO-ST Bearing (PSD) | Domain Adaptation |
| Gribbestad et al. (2021) [44] | Supervised | RUL | Multivar. Time Series | Proprietary Air Compressor; NASA Turbofan | Parameter Transfer |

| Marei *et al.* (2021) [47] | Supervised | RUL | Images | Proprietary Milling 2 | Shared Feature Extractor plus Finetuning |
| Maschler *et al.* (2021) [25] | Supervised | SOH$_{class}$ | (Hand-crafted) Features | NASA Battery | Regularization |
| Zeng *et al.* (2021) [41] | Unsupervised | RUL | Univar. Time Series | FEMTO-ST Bearing | Domain Adaptation |
| Zhang *et al.* (2021) [42] | Supervised | RUL | Multivar. Time Series | NASA Turbofan; XJTU-SY Bearing | Domain Adaptation |

## 4.2 DATA SETS

In order to demonstrate the deep industrial transfer learning algorithms' applicability to real-world problems, the data sets used need to reflect the complexity and dynamics of such problems. Therefore, the data sets used in the surveyed publications are listed in TABLE 3 – with the exception of proprietary data sets only accessible to some researchers and because of that not relevant to a wider audience. Two recently published data sets not yet used in publications were added in order to include them in the discussion in chapter 4. In the following, all listed data sets are described:

For the *NASA Milling* data set [48], sixteen milling tools of unspecified type were run to failure on a MC-510V milling center under eight different operating conditions characterized by different depth of cut, feed rate and material. Acoustic emissions, vibrations and current are recorded with 250 Hz resulting in approximately 1.5 million multivariate samples.

For the *NASA Bearing* data set [49], twelve bearings Rexnord ZA-2115 were run to failure with four being simultaneously on the same shaft but subject to different radial forces. Horizontal and vertical (for some bearings only one) acceleration signals are recorded with 20 kHz for approximately 1 second of every tenth minute, resulting in approximately 15.5 million multivariate (bi- respectively univariate if only one bearing is used) samples. However, because the experiments were stopped once one bearing became faulty, there are exact RUL labels for only four bearings (one experiment encountered a double fault).

For the *NASA Battery* data set [50], 34 type 18650 lithium-ion battery cells were run to failure under 34 different operating conditions characterized by different ambient temperatures, discharge modes and stopping conditions. For each (dis-)charging cycle, static sensor values as well as different time series at 1 Hz, e. g. battery voltage, current and temperature, are recorded. Therefore, there are two levels of multivariate time series contained in this data set. The total number of samples is approximately 7.3 million multivariate samples. However, failures of measurement equipment for some of the batteries as well as other anomalies lead to a lower number of labeled and usable samples [25].

For the *NASA Turbofan* data set [51], 1,416 virtual turbofan engines were run to failure under six different operating conditions characterized by altitude, throttle resolver angle and speed using the so-called Commercial Modular Aero-Propulsion System Simulation (C-MAPSS). 26 different sensor values are recorded as snapshots once per simulated flight resulting in 265,256 multivariate samples.

The *CALCE Battery* data set [52] is an extensive collection of run-to-failure experiments on different types of batteries. The only study included in this survey utilizing some of this data is [45] – we will therefore focus on the "CS2" data set used there. It consists of data from thirteen lithium-ion batteries run to failure under six different operating conditions characterized by different discharge currents and different cut-off voltages. Six different electric sensor values are recorded approximately every 30 seconds resulting in 8.4 million multivariate samples. Two more lithium-ion batteries were also measured, however, the measured characteristics are different from the others and therefore excluded in this overview.

For the *FEMTO-ST* Bearing data set [53], seventeen bearings were run to failure under three different operating conditions characterized by different rotating speeds and different radial force. Horizontal and vertical acceleration signals are recorded with 25.6 kHz 0.1 seconds every 10 seconds and (for only thirteen specimens) temperature signals with 10 Hz, resulting in approximately 63.7 million multivariate samples.

After a multi-year period without the release of major new data sets used in transfer learning studies, recent years finally brought a new wave of larger, more complex data sets:

For the *XJTU-SY* Bearing data set [22], five heavy duty bearings LDK UER204 each were run to failure under three different operating conditions characterized by different rotating speeds and different radial force. Horizontal and vertical acceleration signals are recorded with 25.6 kHz for 1.28 seconds of every minute, resulting in approximately 302 million bivariate samples. Usually, only the horizontal acceleration is used for fault prediction.

To the best of our knowledge, the following two newest data sets have not been used in published transfer learning studies yet. However, due to their complexity, they appear highly suitable for transfer learning evaluation scenarios and should therefore be brought to the community's attention:

The *NASA Turbofan 2* data set [54] features a high number of non-trivial data dimensions and thereby provides a highly complex problem scenario. For this data set, nine virtual turbofan engines were run to failure using C-MAPSS parametrized by real flight data. Seven engines had similar operating conditions, whereas two had individual operating conditions. 45 different sensor values are recorded with 1 Hz resulting in 6.5 million multivariate samples.

The *US Relays* data set [55] features a high number of operating conditions, specimens and samples and thereby provides highly complex problem scenario. For this data set, 100 electromechanical relays of 5 different types were run to failure under 22 different operating conditions characterized by different supply voltages and different load resistances. For each switching cycle, a number of static sensor values as well as voltage time series were recorded. Therefore, there are two levels of multivariate time series contained in this data set. The total number of samples is approximately 162.5 million multivariate samples.

**TABLE 3.**   Overview of fault prognostics data sets used in deep industrial transfer learning publications

| Source | Data Set Name | Data Type(s) | No. of Operating Conditions | No. of Specimens | No. of Samples* |
|---|---|---|---|---|---|
| Agogino et al. (2007) [48] | NASA Milling | Multivariate Time Series | 8 | 16 | 1,503,000 |
| Lee *et al.* (2007) [49] | NASA Bearing | Multivariate** Time Series | 3 | 12 | 15,540,224 |
| Saxena *et al.* (2007) [50] | NASA Battery | Multivariate Time Series | 34 | 34 | 7,282,946 |
| Saxena *et al.* (2008) [51] | NASA Turbofan | Multivariate Time Series | 6 | 1,416 | 265,256 |
| CALCE (2011) [52] | CALCE Battery | Multivariate Time Series | 6 | 13 | 8,438,937 |
| Nectoux *et al.* (2012) [53] | FEMTO-ST Bearing | Multivariate Time Series | 3 | 17 | 63.718.828 |
| Wang *et al.* (2020) [22] | XJTU-SY Bearing | Bivariate Time Series | 3 | 15 | 302,000,000 |
| Arias Chao *et al.* (2021) [54] | NASA Turbofan 2 | Multivariate Time Series | 3 | 9 | 6,500,000 |
| Maschler *et al.* (2022) [55] | US Relays | Multivariate Time Series | 22 | 100 | 162,500,000 |

\*    Samples are defined as individual data tuples measured at respectively representing a different time or using a different measurement object than any other data tuple. If the data set differentiates between test, validation or training data, this differentiation is ignored here and the maximum number of unique samples considered.

\*\*   The time series are multivariate for sets of four bearings and bi- respectively univariate for individual bearings.

## 5. DISCUSSION

In this chapter, the results of the systematic literature review are discussed and further analyzed. First, learnings and best practices are derived from the original research publications in order to consolidate the current state of research in this field. Then, open-access fault prognostics data sets are examined and suitable ones for benchmarking identified.

### *5.1 APPROACHES*

The presented approaches all belong to the **solution categories** of feature representation and parameter transfer of transfer learning and the regularization strategies of continual learning, which, however, methodically represent a form of parameter transfer (see chapter 2.1). Various deep learning methods are utilized, from simple FCNNs and RNNs of various forms to autoencoders or complex, pre-trained CNNs such as AlexNet or ResNet.

As **input data type**, primarily time series data is used, which is typical for industrial applications. Only occasionally are features, image or meta data used directly [25, 43, 45, 47]. Only some of the presented approaches process multivariate time series [27, 32, 33, 37, 42, 44] and no approach uses different types of data in parallel. The complexity of the scenarios used for evaluation is therefore still low, lacking evidence for the applicability of the approaches presented towards more diverse and dynamic real-life scenarios.

Most of the publications presented use only a single data set for **evaluation**. Only [42–45] evaluate their algorithm on different, similar data sets and no study uses evaluation data sets with different data types. Thus, statements about the transferability of the approaches presented are based upon only thin evidence, ready-to-use architectures or frameworks are not available, yet.

Only two of the publications presented address **SoH$_{class}$ estimation** [25, 27] for fault prognostics, which makes it a fringe approach. Direct RUL prediction is much more prominently represented, marking it the default problem category in the field of fault prognostics.

The **results** of some publications are not well enough documented to allow a clear evaluation of the presented approaches' performance. This may be due to a lack of comparative values [34, 41, 47] or their insufficient documentation [40]. [43] appears unfinished due to the generic input data treatment which is not adapted to the use case. In general, although six studies make use of the NASA Turbofan data set and five utilize the FEMTO-ST bearing data set and, thereby, allow benchmarking to some degree, the field would benefit from ubiquitous comparability based upon a common set of benchmarking algorithms and well-documented methodologies as well as evaluation results. This would greatly increase the identification of generalizable best practices and by this speed up scientific progress [1, 4].

**Future research** should build upon the following findings: The appropriate selection of source data to be used in a transfer increases said transfer's performance [30]. Furthermore, [42] shows that a separate treatment of different sample clusters, e.g. of previously known sub-scenarios, increases a transfer's performance as well. Regularization approaches, however, do not appear suitable due to their complex dependencies [25, 27]. Regarding the use of vibration data, utilizing preprocessed data (e.g. FFT) improves results compared to utilizing raw data.

### 5.2 DATA SETS

The data sets used in the presented publications are predominantly open-access, with only four exceptions in [4, 43, 44, 47]. Due to their low accessibility, proprietary data sets obviously are unsuitable for benchmarking purposes and should be accompanied by the use of open-access data sets in any high-impact publication.

So far, the NASA Ames Research Center has provided most of the open-access data sets used for research in the field of industrial transfer learning for fault prognostics – notably, the FEMTO-ST data set is made available through their repository as well. The most widely used data sets were promoted by the Prognostics and Health Management (PHM) challenges of 2008 [51] and 2021 [53]. It is plausible that [54] will encounter a similar effect by the PHM challenge of 2021. Contests like the PHM challenges facilitate comparability of approaches and results by forcing the participants to use the same data sets for evaluation. Furthermore, institutional repositories such as NASA Ames' data repository increase the chance of the contained data sets' long-term availability compared to private or individual university chairs' websites. Both aspects, **wide-spread usage** and **long-term availability**, are qualities required by a benchmark data set.

Apart from those meta-criteria, a benchmark data set should reflect the challenges typical for the respective application or usage scenario. In this case, they should therefore be **heterogenous** and **dynamic**, ideally consisting of (many) different data dimensions, operating conditions and specimen while providing a high number of samples to train and evaluate algorithms on. Even if the sub-problem to be solved by an algorithm does not in itself require the full complexity, using a very complex data set will allow comparability with a wider array of other publications and should therefore be preferable to a data set of lesser complexity. These criteria are best met by the two newest data sets presented in [54, 55].

## 6. CONCLUSION

Fault prognostics is of great importance, e.g. in reducing downtime in manufacturing, harm by failures or wastage by premature maintenance, and deep-learning-based approaches show promising results in research environments. However, in order to be applicable to the heterogenous and dynamic nature of real-world industrial scenarios, deep industrial transfer learning capabilities are required to lower the effort of data collection and algorithm adaptation to new (sub-)problems.

This article introduces transfer learning's main approaches of feature representation and parameter transfer as well as continual learning's regularization strategy. It then describes the basics of fault prognostics regarding different approaches and different objectives. The systematic literature review presents a comprehensive overview of the approaches published on the topic of fault prognostics by deep industrial transfer learning. Despite the diverse array of approaches, results and applications, there is a lack of comparability. Still, some best practices e.g. regarding source data selection and handling can be identified. An ensuing review of open-access data sets for fault prognostics by deep industrial transfer learning underlines the availability of a variety of such data sets. However, only some of them provide the complexity necessary to allow the full range of transfer scenarios – and only those are fully suitable for benchmarking purposes. Luckily, after a few years without new data sets, there have recently been notable publications.

Thereby, this article provides a range of state-of-the-art examples and analyses for anyone considering to enter the field of fault prognostics by deep industrial transfer learning.